\documentclass{ifacconf}
\usepackage{subfig}

\usepackage{multirow}
\usepackage{graphicx}      % include this line if your document contains figures
\usepackage{natbib}        % required for bibliography
\usepackage{bm}            % for bold text in math env

\usepackage{subfloat}

\usepackage{xcolor}
\usepackage{amsmath}
\usepackage{amsfonts}
\usepackage{graphicx} 
\usepackage{url} 
\usepackage{hhline}

\begin{document}
\begin{frontmatter}

\title{Finetuning Deep Reinforcement Learning Policies with Evolutionary Strategies for Control of Underactuated Robots \thanksref{footnoteinfo}  
} 

\thanks[footnoteinfo]{Gian Antonio Susto was supported by PNRR research activities of the consortium iNEST (Interconnected North-East Innovation Ecosystem) funded by the European Union Next GenerationEU (Piano Nazionale di Ripresa e Resilienza (PNRR) – Missione 4 Componente 2, Investimento 1.5 – D.D. 1058 23/06/2022, ECS 00000043). This manuscript reflects only the Authors’ views and opinions, neither the European Union nor the European Commission can be considered responsible for them.
\\Correspondence to calmarco@dei.unipd.it.}

\author[first]{Marco Calì} 
\author[second]{Alberto Sinigaglia} 
\author[first]{Niccolò Turcato}
\author[first]{Ruggero Carli}
\author[first,second]{Gian Antonio Susto}

\address[first]{Department of Information Engineering, University of Padova, Via Gradenigo 6/B, Padova, 35131, Italy}
\address[second]{Human-Inspired Technology Research Center, University of Padova, Via Luzzatti, 4, Padova, 35121, Italy}

\begin{abstract}                % Abstract of 50--100 words
Deep Reinforcement Learning (RL) has emerged as a powerful method for addressing complex control problems, particularly those involving underactuated robotic systems. However, in some cases, policies may require refinement to achieve optimal performance and robustness aligned with specific task objectives. In this paper, we propose an approach for fine-tuning Deep RL policies using Evolutionary Strategies (ES) to enhance control performance for underactuated robots. Our method involves initially training an RL agent with Soft-Actor Critic (SAC) using a surrogate reward function designed to approximate complex specific scoring metrics. We subsequently refine this learned policy through a zero-order optimization step employing the Separable Natural Evolution Strategy (SNES), directly targeting the original score. Experimental evaluations conducted in the context of the 2nd AI Olympics with RealAIGym at IROS 2024 demonstrate that our evolutionary fine-tuning significantly improves agent performance while maintaining high robustness. The resulting controllers outperform established baselines, achieving competitive scores for the competition tasks.
\end{abstract}

\begin{keyword}
Control Systems, Deep Learning, Evolutionary algorithms, Reinforcement learning, Underactuated systems.
% Five to ten keywords, preferably chosen from the IFAC keyword list.
\end{keyword}

\end{frontmatter}
%===============================================================================
\section{Introduction}

Reinforcement Learning (RL) has demonstrated remarkable success in solving complex control problems, including tasks involving underactuated robotic systems \citep{deepRLsurvey,WANG2022100066}. However, policies often require further refinement to align with specific task objectives. 
Specifically, several robotics tasks are evaluated by computing sparse scoring metrics based on task requirements and objectives, e.g., task completion time, maximum energy, maximum velocity, maximum action, or a combination of them.
Typically, to optimize scores or evaluation metrics based on that kind of criteria, RL agents are trained on surrogate dense reward functions, especially in continuous control settings that benefit from dense rewards to guide exploration. This approach, although effective, inevitably creates biased agents, depending on the quality of the surrogate function. For example, swing-up tasks in underactuated robots are typically addressed using distance-based punctual costs or rewards, which are surrogates for the swing-up time and similar criteria, e.g. \cite{mcpilco, competition-1st-edition, turcato2024learningcontrol, black-drops}.
In this paper, we present an approach to fine-tune policies learned via Deep RL with evolutionary strategies \citep{Back1996ES}, to optimize general sparse scoring metrics.
In this work, we focus on Soft Actor-Critic (SAC) \citep{sac} as the reference model-free RL algorithm. 
Therefore, we refer to the proposed algorithm as \emph{Evolutionary SAC} (EvolSAC). Still, the methodology can be extended to other model-free and model-based RL algorithms.
\\
Initially, we train an RL agent using SAC with a comprehensive reward function that captures the fundamental aspects of the task, such as the desired configuration, and a penalty on the kinetic energy and torques. SAC provides a stable and sample-efficient training process compared to other MFRL alternatives, allowing us to approximate the original score and efficiently achieve strong baseline performance. Furthermore, SAC's entropy-regularized optimization framework naturally encourages robust policies that generalize well to different conditions.
We introduce an additional optimization step leveraging evolutionary strategies to further refine the policy and optimize for the real score function. This fine-tuning phase aligns the agent’s behavior more closely with the desired objective, mitigating discrepancies between the surrogate reward function used during RL training and the true evaluation metric.

\begin{figure*}[t]
    \centering
    \includegraphics[width=0.7\linewidth]{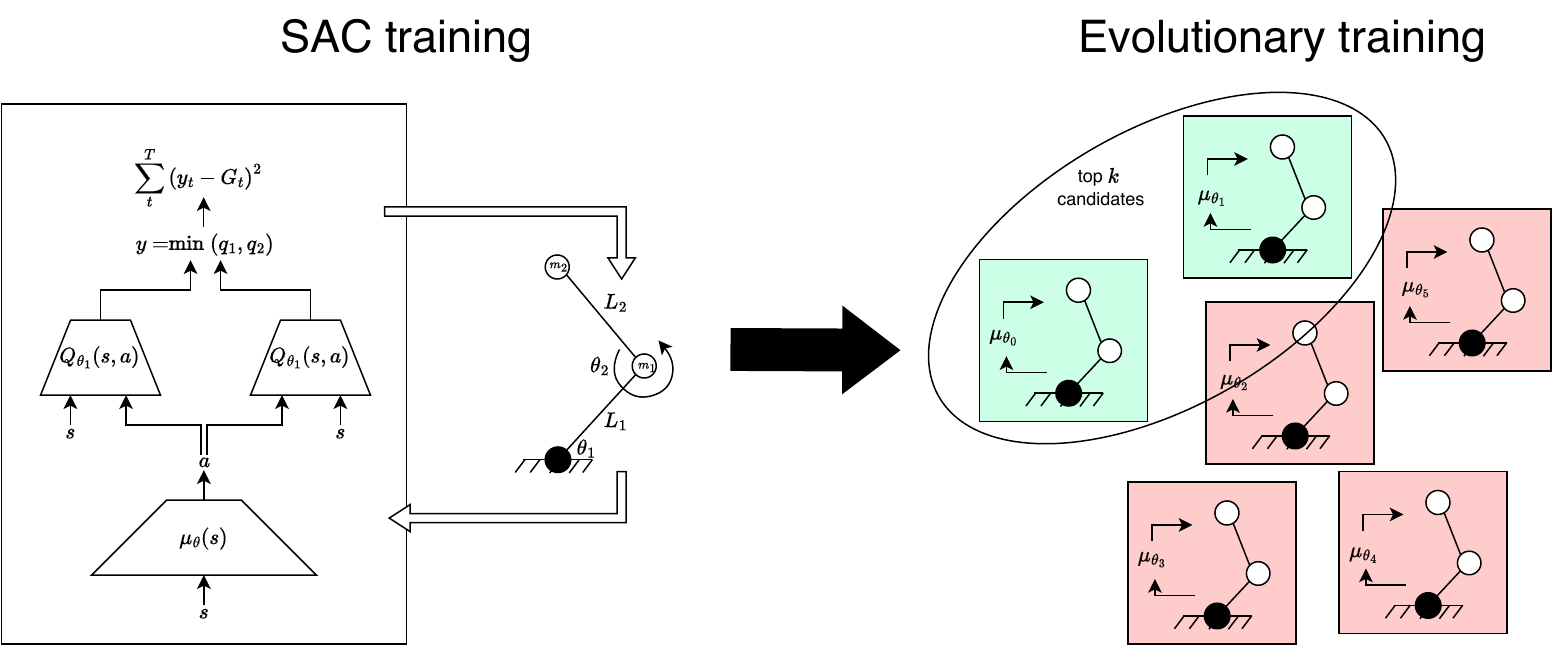}
    \caption{Schematics of the training procedure of Evolutionary SAC. First, the agent is trained to maximize the surrogate reward function. Then, the optimal policy from the SAC training is optimized via evolutionary strategies.}
    \label{fig:evolsac}
\end{figure*}

We first evaluate our approach in simplified settings, such as the classic cartpole task, to demonstrate its basic effectiveness. We then validate its performance on a more challenging, competition-level benchmark: the ``2nd AI Olympics with RealAIGym at IROS 2024'' \citep{competition-1st-edition,wiebe2025reinforcementlearningrobustathletic}. This benchmark focuses on the swing-up and stabilization of an underactuated double pendulum, tested in both acrobot and pendubot configurations, around their unstable upright equilibria.

\section{Background}\label{sec:background}

RL involves training an agent to make decisions by interacting with an environment, aiming to maximize cumulative rewards.
RL is typically modeled using Markov Decision Processes (MDPs), defined by the tuple $(S, A, P, R, \gamma)$, where $S$ is the set of states, $A$ is the set of actions, $P$ defines the transition probabilities $P(s'|s,a)$. $R$ is the reward, a function of state, action, and next state $R(s,a,s')$, $\gamma$ is a discount factor. The return $G_t$ at time $t$ is defined as:
\begin{equation}
    G_t = \sum_{k=0}^{\infty} \gamma^k R_{t+k+1}.
\end{equation}

The value function $V(s)$ estimates the expected return from state $s$:
\begin{equation}
V(s) = \mathbb{E}[G_t | s_t = s].
\end{equation}

The action-value function $Q(s,a)$ estimates the expected return after taking action $a$ in state $s$:
\begin{equation}
Q(s,a) = \mathbb{E}[G_t | s_t = s, a_t = a].
\end{equation}

Finally, a \textit{policy} $\pi$ is a mapping from states to actions or probabilities of actions, defined formally as $\pi(a|s)$, representing the probability of selecting the action $a$ given the state $s$. The goal of RL is typically to find an optimal policy that maximizes the expected cumulative reward.

\subsection{Deep Deterministic Policy Gradient (DDPG)}
The Deep Deterministic Policy Gradient (DDPG) algorithm \citep{ddpg} extends the Deterministic Policy Gradient (DPG) framework \citep{dpg} to leverage deep neural networks for function approximation in continuous action spaces. DDPG maintains an actor-critic architecture, where the actor \(\pi_\theta(s)\) deterministically maps states to actions, and the critic \(Q_\phi(s, a)\) estimates the action-value function associated with the current policy. The objective for the actor is to maximize the expected return under the learned Q-function:
\begin{equation}\label{eq:ddpg_objective}
    J(\pi_\theta) = \mathbb{E}_{s_t \sim \rho} \left[ Q_{\phi}(s_t, \pi_\theta(s_t)) \right],
\end{equation}
where $\rho$ is the state distribution induced by the policy. 
\\
Training proceeds by sampling transitions \((s, a, r, s')\) from a replay buffer, which helps break correlations between sequential data and improves sample efficiency. The critic is updated by minimizing a temporal difference loss with respect to the target Q-value:
\begin{equation}\label{eq:temporal_difference_loss}
    \mathcal{L}(\phi) = \left( r + \gamma Q_{\phi'}(s', \pi_{\theta'}(s')) - Q_\phi(s, a) \right)^2,
\end{equation}
where \(\phi'\) and \(\theta'\) denote the parameters of slowly updated target networks that stabilize training by decoupling the target from the learned Q-function and policy.
\\
The actor is updated by following the gradient of the Q-function with respect to the action, backpropagated through the actor parameters:
\begin{equation}
    \nabla_\theta J(\pi_\theta) = \mathbb{E}_{s \sim \mathcal{D}} \left[ \nabla_a Q_\phi(s, a) \big|_{a = \pi_\theta(s)} \nabla_\theta \pi_\theta(s) \right],
\end{equation}
where \(\mathcal{D}\) denotes the replay buffer distribution.
\\
By combining deterministic policy gradients with deep function approximators, target networks, and experience replay, DDPG enables stable learning in high-dimensional continuous control problems. However, the algorithm is sensitive to hyperparameters and may suffer from overestimation bias and premature convergence.

\subsection{Twin Delayed Deep Deterministic Policy Gradient (TD3)}
The Twin Delayed Deep Deterministic Policy Gradient (TD3) algorithm \citep{td3} improves upon DDPG by addressing critical limitations such as overestimation bias and training instability, which are common in actor-critic methods. TD3 introduces three key enhancements to improve performance and robustness.
\\
First, TD3 employs clipped double Q-learning, in which two independent critic networks are trained simultaneously, and the minimum of their Q-value estimates is used during the target computation. This conservative estimate mitigates over-optimistic value predictions that can mislead the actor. Second, delayed policy updates are introduced: the actor and target networks are updated less frequently than the critics, allowing the Q-function to converge more reliably before influencing the policy. Lastly, target policy smoothing adds noise to the target action during the critic update. This  smoothes out sharp peaks in the Q-function, reducing the likelihood that the actor exploits narrow value spikes.

\subsection{Soft Actor-Critic (SAC)}
The Soft Actor-Critic (SAC) algorithm \citep{sac} introduces an entropy term to the reward structure, promoting exploration and robustness in policy learning. SAC optimizes a stochastic policy to balance the trade-off between entropy and reward. The objective function for SAC is given by:
\begin{equation}\label{eq:sac_objective}
    J(\pi_\phi) = \sum_{t=0}^T \mathbb{E}_{s_t, a_t \sim \rho_\pi} \left[ r(s_t, a_t) + \alpha \mathcal{H}(\pi_\phi(\cdot | s_t)) \right)].
\end{equation}

Here, \(\mathcal{H}\) denotes the entropy of the policy \(\pi_\phi\), where $\phi$ is the set of parameters of the network, and \(\alpha\) is a temperature parameter that controls the importance of the entropy term against the reward. SAC employs separate networks for the policy, value function, and two Q-functions, reducing overestimation bias and improving learning stability; $\rho_\pi$ is the state-action marginal of the trajectory distribution induced by the policy $\pi_\phi$. 

\subsection{Zero-Order Optimization and Evolutionary Strategies}
Zero-order optimization, or derivative-free optimization, encompasses a range of techniques that optimize functions without requiring gradient information. These methods are essential in scenarios where gradients are non-existent, expensive to compute, or do not reliably guide the optimization due to noise, discontinuities, or non-differentiability. Evolutionary strategies (ES) are a prominent class of zero-order optimization techniques that employ mechanisms analogous to biological evolution, such as mutation, recombination, and selection, to evolve a population of candidate solutions over successive generations. However, they heavily rely on large-scale computation to achieve effectiveness, which might be extremely limiting for their application in robotic scenarios.
   
\subsection{Separable Natural Evolution Strategy (SNES)}
The Separable Natural Evolution Strategy (SNES) \citep{snes} refines evolutionary strategies by focusing on the efficient adaptation of mutation distributions. SNES is particularly effective in environments where parameters have different scales and sensitivities. It maintains and adapts a separate step size for each parameter dimension, facilitating an independent and efficient exploration of the parameter space. The core of the SNES algorithm involves updating the mutation strengths using a log-normal rule, which is mathematically depicted as follows:
\begin{align}
    \sigma_{\text{new},i} &= \sigma_{\text{old},i} \exp \left( \tau \mathcal{N}(0, 1) + \tau' \mathcal{N}(0, 1)_i \right), \\
    \theta_{\text{new},i} &= \theta_{\text{old},i} + \sigma_{\text{new},i} \mathcal{N}(0, 1)_i,
\end{align}
for each dimension \(i\), where \(\tau\) and \(\tau'\) are learning rates designed to control the global and individual adaptation speeds, respectively. This adaptation mechanism is inspired by the principles of natural evolution and covariance matrix adaptation, but it simplifies the adaptation process by treating the search space dimensions as separate entities. The SNES thus combines the robustness of evolutionary algorithms with the efficiency of adaptive step-size mechanisms, leading to faster convergence and reduced computational complexity compared to traditional ES and other sophisticated adaptation strategies like CMA-ES \citep{hansen2001completely}.
\section{Proposed Approach}\label{sec:proposed_approach}
RL approaches typically define the reward function as $R(s, a, s')$, specifying rewards based solely on the current state transitions. Although this formulation is sufficiently general for numerous applications, it can prove limiting in scenarios where rewards depend explicitly on future events or global properties of the trajectory. A pertinent example occurs when the reward function includes terms such as the maximum action exerted over an entire trajectory or task completion time, which inherently depend on future states and actions. In contrast, evolutionary strategies (ES) are naturally suited for handling such reward functions, as they operate by directly optimizing policy parameters based on cumulative or trajectory-wide metrics. However, ES approaches are notoriously data-inefficient, often requiring substantially more interactions with the environment to achieve convergence.
\\
To address these challenges, we propose a hybrid two-phase training methodology. The (i) first phase employs Deep RL to rapidly identify a policy capable of effectively solving the task using a surrogate reward function. This surrogate reward is designed to correlate with the true objective while being easier to optimize. 
In the (ii) second phase, we employ evolutionary strategies to fine-tune the policy parameters, explicitly optimizing for the true objective and bridging the gap between the surrogate reward and the actual target criterion. Leveraging the initial policy obtained via RL as a starting point significantly enhances the efficiency of the evolutionary optimization step, enabling rapid convergence to a solution that effectively optimizes complex trajectory-dependent reward criteria. Figure~\ref{fig:evolsac} shows a high-level representation of the approach.

The SAC framework encompasses three primary neural networks: a policy network and two Q-networks. These networks are parametric models; hence, they can be optimized through zero-order methods such as evolutionary strategies. Such strategies excel in optimizing systems that are either highly ill-conditioned or non-differentiable. Subtle modifications to the policy parameters can yield considerable disparities in the behavior of controllers, consequently affecting performance metrics drastically. The optimization of SAC is inherently non-stationary, as the policy updates are executed through Stochastic Gradient Descent (SGD) on the evolving Q-functions.
Notably, SNES requires minimal hyperparameter tuning, with the key parameters being the variance $\sigma$ of the population and the population size.

\begin{figure}[h]
    \centering
    \includegraphics[width=\linewidth]{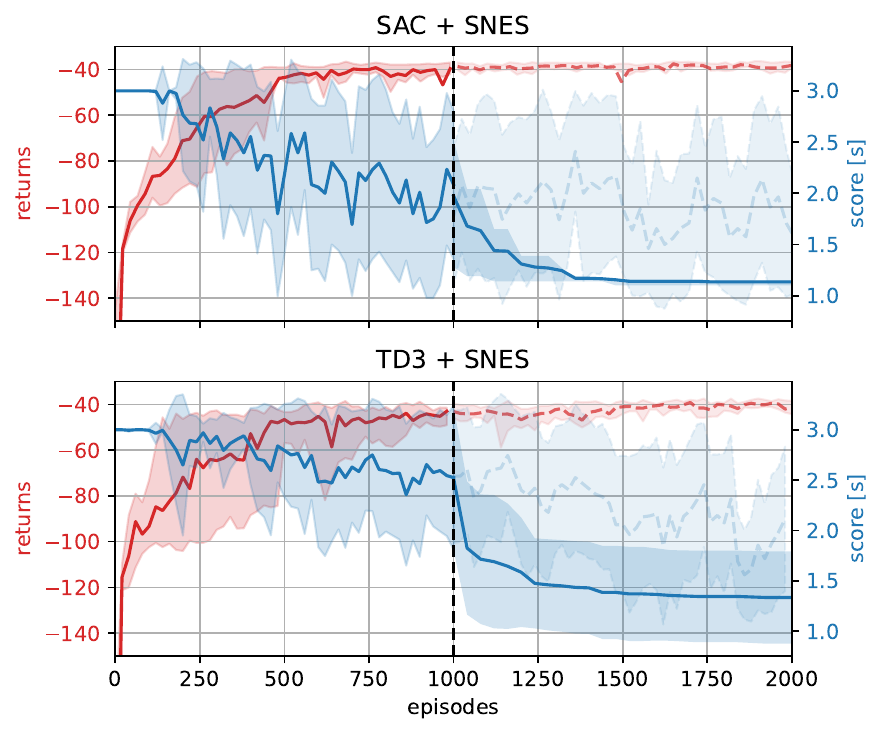}
    \caption{Comparison between SAC + SNES and TD3 + SNES on the cartpole swing-up task.}
    \label{fig:comparison_between_sac_td3_plus_snes} 
\end{figure}

\subsubsection{On the agent robustness}
As SAC optimizes eq. ~\eqref{eq:sac_objective}, its agent models $\pi_\phi(s) \sim \mathcal N(\mu(s), \text{diag}[\Sigma(s)])$. Thanks to this definition, the model is already optimized to find a reasonably robust solution. Indeed, the agent has to integrate all actions that it can take in the next state, which are normally distributed, and it's optimized to reward high-entropy distributions. For this reason, the solution found by SAC will already be robust to noise on the action sampled.
\\
The SNES algorithm is employed to align the agent’s behavior with the actual reward function rather than the surrogate one. However, optimizing this reward directly can introduce a significant drawback: while SAC tends to produce agents that are robust by design, SNES lacks any inherent mechanism to prevent overfitting to a single, highly effective—but potentially brittle—trajectory. As a result, the agent may learn to exploit specific conditions that maximize the reward, at the cost of general robustness. For this reason, the training is modified as follows. Each timestep:
\begin{enumerate}
 \item Sample the greedy action from the SAC agent: ${a := \pi_\theta(s)}$;
 \item Undo the $\text{tanh}$ transformation $a := \text{atanh}(a)$;
 \item Introduce noise: $a := a + \epsilon, \,\, \epsilon \sim N(0, \sigma^2)$;
 \item Apply squashing: $a := \text{tanh}(a)$.
\end{enumerate}
The procedure described above prevents the posterior from collapsing to a Dirac delta distribution. The noise variance $\sigma^2$ is obtained by measuring the average variance during a trajectory obtained after training with SAC.

\section{Experimental Results}
In this section, we present the experiments performed on two different underactuated systems typically used as benchmarks, namely the cartpole and the double pendulum in the acrobot and pendubot configurations. 

\subsection{Cartpole}

The cartpole system consists of a pole attached to a cart that moves along a horizontal track. The system is controlled by applying horizontal forces to the cart, with the goal of swinging the pole up and balancing it in the upright position. Let the state be defined as:
\begin{equation}\label{eq:cartpole_state}
    \bm{x}_t = \left[\bm{q}_t;  \dot{\bm{q}}_t\right],
\end{equation}
with $\bm{q}_t = \left[x_t ;  \theta_t\right]$, where $x_t$ and $\theta_t$ are, respectively, the cart position and pole angle at time $t$. Here, $\dot{\bm{q}}_t = [ \dot{x}_t ; \dot{\theta}_t ]$ represents the cart velocity and the angular velocity of the pole, respectively.

The desired target configuration is $\bm{x}_g = \left[0 ; \pm\pi ; 0 ; 0\right]$.

\subsubsection{Swing-up Time Score and Surrogate Reward Function}
An example of a score function that is challenging to optimize directly through a RL approach is the swing-up time, defined in this case as the earliest timestep such that the pole tip remains within a 0.1-unit radius of the upright position for the rest of the episode:
\[
    t_s = \min \left\{ \tau \in [0, T] \mid \left\| \bm{p}_t - \bm{p}_\text{upright} \right\| < 0.1 \quad \forall t \geq \tau \right\},
\]
where the tip position is $\bm{p}_t = \left[x_t-\ell\sin\theta_t;\ell\cos\theta_t \right]$,
and $\bm{p}_\text{upright} = \left[ 0; \ell\right]$,
with $\ell$ being the pole length. To train policies using RL, we use a surrogate reward that penalizes the Euclidean distance between the pole tip and the upright position:
\begin{equation}\label{eq:cartpole_reward}
  R(\bm {x}_t) = -\left\| \bm{p}_t - \bm{p}_\text{upright} \right\|.
\end{equation}
The episode duration is \( T = 3\,\text{s} \), with a timestep of \( 0.01\,\text{s} \) in training, and \(0.05\,\text{s}\) in evaluation. The maximum control input is limited to \( u_{\max} = 2.5\,\text{N} \).

\subsubsection{Training Setup and Results}  
We train policies using SAC and TD3 for 2000 episodes using the reward in \eqref{eq:cartpole_reward}. After this stage, we apply SNES to finetune the policy parameters, starting from the policies obtained at episode 1000, to compare the advantages of the evolutionary fine-tuning. The SNES population size is 40, and the population variance is set to 0.02. During SNES fine-tuning, the policy is evaluated with a timestep of $0.05 \, \text{s}$. We train 10 independent agents.
Figure~\ref{fig:comparison_between_sac_td3_plus_snes} compares the performance of SAC and TD3 before and after fine-tuning with SNES. Initially, the average swing-up times are $2.085\,\text{s}$ for SAC and $2.525\,\text{s}$ for TD3. After fine-tuning with SNES, both methods exhibit significant improvements: SAC+SNES achieves an average swing-up time of $1.135\,\text{s}$ (a $45.6\%$ reduction), while TD3+SNES reaches $1.345\,\text{s}$ (a $46.7\%$ reduction). 
In contrast, continuing training with SAC or TD3 alone for an additional 1000 episodes results in higher swing-up times of $1.565\,\text{s}$ and $2.115\,\text{s}$, respectively, indicating diminishing returns without the evolutionary fine-tuning. Notably, SNES alone does not discover viable solutions within this time frame. As training progresses, both SAC+SNES and TD3+SNES asymptotically approach the swing-up time of $1.12\,\text{s}$.

\subsection{Double Pendulum}
The double pendulum is a planar robotic system with two degrees of freedom (2-DOF), consisting of two rigid links connected in series. The first link is attached to a fixed pivot, and the second is connected to the end of the first. The system is underactuated, with control applied either at the first joint (pendubot) or at the second joint (acrobot).
The state is defined similarly to \eqref{eq:cartpole_state}, with:
\begin{equation}\label{eq:double_pendulum_q_t}
    \bm{q}_t = \left[ \theta_{1,t} ; \theta_{2,t} \right],
\end{equation}
where $\theta_{1,t}$ is the angle of the first link from the vertical and $\theta_{2,t}$ is the relative angle of the second link with respect to the first. Here, $\dot{\bm{q}}_t = [ \dot{\theta}_{1,t} ; \dot{\theta}_{2,t} ]$ contains the angular velocities of the first and second links.
The desired configuration for the system is given by $\bm{x}_g = \left[ \pm \pi ; 0 ; 0 ; 0 \right]$.

\subsubsection{Surrogate reward function}
As described in Section~\ref{sec:proposed_approach}, the setting will deal with two objective functions: the RealAIGym AI Olympics performance score and a surrogate reward counterpart.
The end goal is the competition performance score function, which is defined as:
\begin{align}\label{eq:competition_score}
    S = c_\text{succ} \cdot\left( 1 - \frac{1}{5} \sum_{i=1}^{5} \tanh \left( \pi \frac{x_i}{k_i} \right) \right),
\end{align}
where $x_i$ are the criteria, $k_i$ are scaling constants, both defined in \cite{wiebe2025reinforcementlearningrobustathletic}. 
\\
The competition score function is inherently sparse, as it depends on a binary success variable \( c_{\text{succ}} \in \{0, 1\} \). This makes it difficult to optimize directly, especially since it incorporates trajectory-level information that conflicts with the conventional RL reward formulation \( R(s, a, s') \) unless the entire trajectory is encoded within the state. To address this, we design a surrogate reward function that remains aligned with the competition objective while being more practical for standard RL algorithms. The proposed reward function is defined as follows:

\begin{equation}\label{eq:reward_double_pendulum}
R(\bm{x}_t, a_t) = 
\begin{cases}
\begin{aligned}
    V_t
    &+ \alpha [1+ \cos(\theta_{2,t})]^2 \\
    &- \beta T_t - \rho_1 a_t^2 - \phi_1 \Delta a_t
\end{aligned}
& \text{if } y_t > y_{th} \\
\\
\begin{aligned}
    V_t
    &- \rho_2 a_t^2 - \phi_2 \Delta a_t - \eta \lVert \dot{\bm{q}_t} \rVert^2
\end{aligned}
& \text{otherwise}
\end{cases}
\end{equation}

Here, the action \( a_t \in [-1, 1] \) represents a normalized torque input, and \( \Delta a_t \) captures the change in action between successive time steps, serving to penalize abrupt variations and encourage smoother control. The terms \( V_t \) and \( T_t \) denote the potential and kinetic energy of the system at time \( t \), respectively, while \( \|\dot{\bm{q}_t}\|^2 \) acts as a regularizer against excessively fast movements.
\\
To encourage the two links to align vertically in the upright configuration, the cost includes an alignment term \( [1 + \cos(\theta_{2,t})]^2 \), which is minimized when the second joint angle is near zero. The vertical position of the end-effector is denoted by \( y_t \), with \( y_{\text{max}} = 0.5 \, \text{m} \) representing the maximum achievable height. A system-specific threshold \( y_{\text{th}} \) defines when the higher-reward regime is activated—set to \( 0.375 \, \text{m} \) for the acrobot and \( 0.35 \, \text{m} \) for the pendubot.
\\
Finally, the learning rate is set to $0.001$, and the weights \(\alpha, \beta, \rho_1, \rho_2, \phi_1, \phi_2, \eta \) are tunable hyperparameters that balance the various objectives of the cost function, including energy shaping, control smoothness, and effort minimization. The full set of hyperparameters used in our experiments is provided in Table~\ref{tab:sac_hyperparams}.

\begin{table}[h]
\begin{center}
\begin{tabular}{|c|c|c|c|c|c|c|c|c|}
\hline
$\tau_{\max}$ & $\alpha$ & $\beta$ & $\rho_1$ & $\rho_2$ & $\phi_1$ & $\phi_2$ & $\eta$  \\
\hline
3 & 2 & 1 & 0.1 & 0.02 & 0.15 & 0.15 & 0.02 \\
\hline
\end{tabular}
\end{center}
\vspace{2mm}
\caption{Hyperparameters used for SAC training.}
\label{tab:sac_hyperparams}
\end{table}

Moreover, the performance score is averaged with the robustness score, as described in \cite{wiebe2025reinforcementlearningrobustathletic}, which evaluates the controllers' robustness against changes in the parameters of the system.

\subsubsection{SAC agent training}
The SAC agent is trained to maximize \eqref{eq:reward_double_pendulum}, with the torque limit $\tau_\text{max}$ playing a critical role. Lower torque, e.g. $1.5 \, \text{N} \cdot \text{m}$, encouraged energy-efficient behaviors but led to suboptimal stabilization at configurations like $(\pi, \pi)$ or $(0, \pi)$. Higher torque, e.g. $5.0 \, \text{N} \cdot \text{m}$, accelerated early learning but resulted in excessive and inefficient motions. A balanced setting of $\tau_\text{max} = 3.0 \, \text{N} \cdot \text{m}$ proved optimal, enabling stabilization under $2\, \text{s}$ and efficient control. This configuration achieved top performance scores of $0.504$ for the acrobot and $0.567$ for the pendubot.

\subsubsection{SNES agent training}
Employing evolutionary strategies directly from the competition score proves challenging due to the sparse score function; therefore, the SNES training phase begins from the policy learned during SAC training. Such policy is finetuned with SNES directly on the competition score~\eqref{eq:competition_score}. SNES effectively bridges the gap between the surrogate reward function utilized in training the SAC agent and the original score metric from the competition. Despite its advantages, this methodology is highly contingent on the starting point of the optimization. The resultant policy tends to remain proximate to that of the initial SAC controller, thereby constraining its potential for variation. Nevertheless, we assume that the initial SAC agent has approximated a near-optimal policy, with a discrepancy amenable to further optimization via SNES. The population size for SNES is set to 40, and the variance of the population is set to $0.01$.
\\
Table~\ref{tab:agents_mean_std} presents the mean and standard deviation of performance scores for 5 agents, each trained via SAC with a specific set of hyperparameters, followed by SNES, and evaluated for performance. 
\begin{table}[h]
    \centering
    \begin{tabular}{|c|c|c|}
    \hline
         \textbf{Controller} & \textbf{Acrobot} & \textbf{Pendubot} \\ \hline
        SAC & $0.4972 \pm 0.0088$ & $0.5207 \pm 0.0235$ \\
        EvolSAC & $\mathbf{0.5249 \pm 0.0085}$ & $\mathbf{0.5441 \pm 0.0288}$ \\
        \hline
    \end{tabular}
    \vspace{2mm}
    \caption{Mean and standard deviation of performance score over five trained agents}
    \label{tab:agents_mean_std}
\end{table}

\begin{figure}[thpb]
      \centering
      \includegraphics[width=1\columnwidth]{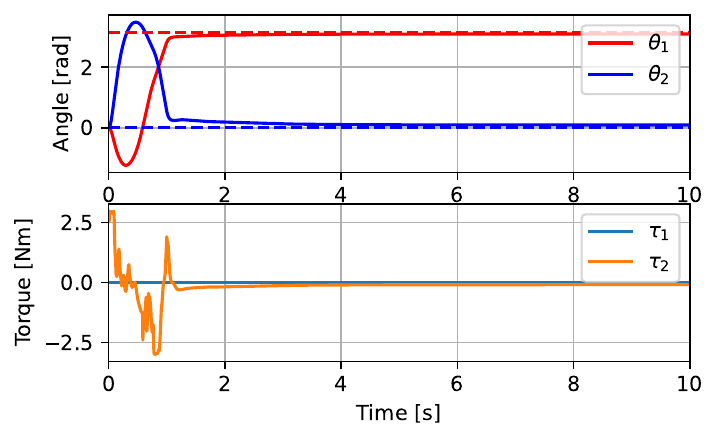}
      \caption{State and input trajectories of the EvolSAC controller - acrobot}
      \label{acrobottrajectory}
\end{figure}

\begin{figure}[thpb]
      \centering
      \includegraphics[width=1\columnwidth]{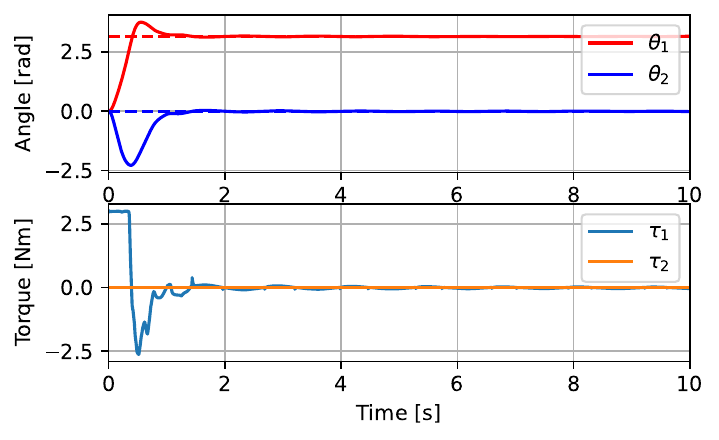}
      \caption{State and input trajectories of the EvolSAC controller - pendubot}
      \label{pendubottrajectory}
\end{figure}

The trajectories resulting from the learned policies are shown in Figures~\ref{acrobottrajectory} and~\ref{pendubottrajectory}.

The optimal agents identified through SNES fine-tuning attain a score of $0.524$ for the acrobot and $0.596$ for the pendubot, as shown in Table~\ref{tab:robustness_comparison_acrobot} and Table~\ref{tab:robustness_comparison_pendubot}. This occurs with a minimal reduction in robustness scores, which drop from $0.700$ to $0.692$ for the acrobot and from $0.800$ to $0.796$ for the pendubot, where by optimal we refer to the highest average score by considering both performance and robustness metrics.
\\
The best results of each controller for acrobot and pendubot settings are compared to the competition baselines \citep{wiebe2025reinforcementlearningrobustathletic} in Table~\ref{tab:robustness_comparison_acrobot} and Table~\ref{tab:robustness_comparison_pendubot}, respectively. EvolSAC exceeds the baseline algorithms in both performance scores and robustness across both system configurations, despite not being explicitly trained to maximize robustness scores.

% \begin{table}[h!]
% \centering
% \begin{tabular}{|c|c|c|c|}
% \hline
% \textbf{Controller} & \textbf{Robustness} & \textbf{Performance} &\textbf{Avg.}\\
% \hline
% TVLQR & 0.607 & 0.504 & 0.555\\
% \hline
% iLQR MPC & 0.343 &0.345& 0.343\\
% \hline
% iLQR Riccati Gains & 0.138 &0.396& 0.267\\
% \hline
% MC-PILCO & 0.241 & 0.316 & 0.279 \\
% \hline
% SAC & 0.700 & 0.504 & 0.602\\
% \hhline{|=|=|=|=|}
% \textbf{Evolutionary SAC} & \textbf{0.692} & \textbf{0.524} & \textbf{0.608}\\
% \hline
% \end{tabular}
% \vspace{2mm}
% \caption{Comparison with baselines - acrobot} \label{tab:robustness_comparison_acrobot}
% \end{table}

% \begin{table}[h!]
% \centering
% \begin{tabular}{|c | c | c |c|}
% \hline
% \textbf{Controller} & \textbf{Robustness} & \textbf{Performance} &\textbf{Avg.}\\
% \hline
% TVLQR & 0.767 & 0.526 & 0.646\\
% \hline
% iLQR MPC & 0.674 & 0.353&  0.513\\
% \hline
% iLQR Riccati Gains & 0.255 & 0.536 & 0.395\\
% \hline
% MC-PILCO & 0.614 & 0.480 & 0.547 \\
% \hline
% SAC & 0.800 & 0.567 & 0.684\\
% \hhline{|=|=|=|=|}
% \textbf{Evolutionary SAC} & \textbf{0.796} & \textbf{0.596} & \textbf{0.696}\\
% \hline
% \end{tabular}
% \vspace{2mm}
% \caption{Comparison with baselines - pendubot} \label{tab:robustness_comparison_pendubot}
% \end{table}

\begin{table}[h!]
\centering
\begin{tabular}{|c|c|c|c|}
\hline
\textbf{Controller} & \textbf{Performance Score} &\textbf{Final Score}\\
\hline
iLQR Riccati Gains & 0.396& 0.267\\
\hline
MC-PILCO & 0.316 & 0.279 \\
\hline
iLQR MPC & 0.345& 0.343\\
\hline
TVLQR & 0.504 & 0.555\\
\hline
SAC & 0.504 & 0.602\\
\hhline{|=|=|=|=|}
\textbf{Evolutionary SAC} & \textbf{0.524} & \textbf{0.608}\\
\hline
\end{tabular}
\vspace{2mm}
\caption{Comparison with baselines - acrobot} \label{tab:robustness_comparison_acrobot}
\end{table}

\begin{table}[h!]
\centering
\begin{tabular}{|c | c | c |c|}
\hline
\textbf{Controller} & \textbf{Performance Score} &\textbf{Final Score}\\
\hline
iLQR Riccati Gains& 0.536 & 0.395\\
\hline
iLQR MPC & 0.353&  0.513\\
\hline
MC-PILCO & 0.480 & 0.547 \\
\hline
TVLQR & 0.526 & 0.646\\
\hline
SAC & 0.567 & 0.684\\
\hhline{|=|=|=|=|}
\textbf{Evolutionary SAC} & \textbf{0.596} & \textbf{0.696}\\
\hline
\end{tabular}
\vspace{2mm}
\caption{Comparison with baselines - pendubot} \label{tab:robustness_comparison_pendubot}
\end{table}

\section{Conclusions}
This work investigates continuous control scenarios where the true objective is not directly optimizable through standard reinforcement learning methods. To tackle this, we first train policies using deep RL algorithms, such as SAC, on a surrogate reward function that is feasible to optimize. We then refine these policies using evolutionary strategies—in particular, SNES—which directly target the true objective, such as minimizing swing-up time or energy usage. This hybrid pipeline combines the sample efficiency of RL with the flexibility of black-box optimization. We evaluate our method on two systems of increasing complexity: the classic cartpole swing-up task and the AI Olympics challenge, featuring an underactuated double-pendulum. Results demonstrate improved performance and robustness over both standard RL and classical control baselines.

% \begin{ack}
% \end{ack}

\bibliography{ifacconf}             % bib file to produce the bibliography
                                                     % with bibtex (preferred)
                                                   
%\begin{thebibliography}{xx}  % you can also add the bibliography by hand

%\bibitem[Able(1956)]{Abl:56}
%B.C. Able.
%\newblock Nucleic acid content of microscope.
%\newblock \emph{Nature}, 135:\penalty0 7--9, 1956.

%\bibitem[Able et~al.(1954)Able, Tagg, and Rush]{AbTaRu:54}
%B.C. Able, R.A. Tagg, and M.~Rush.
%\newblock Enzyme-catalyzed cellular transanimations.
%\newblock In A.F. Round, editor, \emph{Advances in Enzymology}, volume~2, pages
%  125--247. Academic Press, New York, 3rd edition, 1954.

%\bibitem[Keohane(1958)]{Keo:58}
%\newblock \emph{Power and Interdependence: World Politics in Transitions}.
%\newblock Little, Brown \& Co., Boston, 1958.

%\bibitem[Powers(1985)]{Pow:85}
%\newblock Is there a way out?
%\newblock \emph{Harpers}, pages 35--47, June 1985.

%\bibitem[Soukhanov(1992)]{Heritage:92}
%\newblock \emph{{The American Heritage. Dictionary of the American Language}}.
%\newblock Houghton Mifflin Company, 1992.

%\end{thebibliography}

% \appendix
% \section{A summary of Latin grammar}    % Each appendix must have a short title.
% \section{Some Latin vocabulary}              % Sections and subsections are supported  
%                                                                          % in the appendices.
\end{document}